\pdfoutput=1
\documentclass[conference]{IEEEtran}

\usepackage[T1]{fontenc}
\usepackage[utf8]{inputenc}
\usepackage[english]{babel}
\usepackage{cite}
\usepackage{amsmath,amssymb}
\usepackage{graphicx}
\usepackage{booktabs}
\usepackage{tabularx}
\usepackage{array}
\usepackage{enumitem}
\usepackage{listings}
\usepackage{xcolor}
\usepackage{xurl}
\usepackage[hidelinks]{hyperref}

\graphicspath{{figures/}}

\begin{document}

\title{CommandSwarm: Safety-Aware Natural Language-to-Behavior-Tree Generation for Robotic Swarms}

\author{
\IEEEauthorblockN{Mohammed Majeed}
\IEEEauthorblockA{
SlimX.ai\\
\texttt{Mohammed.Majeed@slimx.ai}
}
\and
\IEEEauthorblockN{Amjad Yousef Majid}
\IEEEauthorblockA{
SlimX.ai\\
\texttt{Amjad.Majid@slimx.ai}
}
}

\maketitle

\begin{abstract}
Natural-language interfaces can make swarm robotics more accessible to non-expert operators, but they must translate ambiguous user intent into executable swarm behaviors without unsupported actions, malformed programs, or unsafe plans. This paper presents \emph{CommandSwarm}, a safety-aware language-to-behavior-tree pipeline for generating XML behavior trees (BTs) from speech or text commands. The system combines multilingual translation, command-level safety filtering, constrained prompting, a LoRA-adapted large language model (LLM), and deterministic parser validation against a whitelist of executable swarm primitives. We evaluate eleven open 6.7B--14B parameter LLMs, all using 4-bit quantization, on representative swarm-control scenarios under zero-shot, one-shot, and two-shot prompting. Falcon3-Instruct-10B and Mistral-7B-v3 are the strongest prompt-engineered candidates, reaching BLEU scores above 0.60 and high syntactic validity in few-shot settings. LoRA adaptation of Falcon3-Instruct-10B on a 2,063-example synthetic instruction--BT corpus improves zero-shot BLEU from 0.267 to 0.663, ROUGE-L from 0.366 to 0.692, and parser-accepted syntactic validity from 0\% to 72\%. Translation experiments further show that SeamlessM4T v2-large and EuroLLM-9B provide the best quality-latency trade-offs for the multilingual front end. The results indicate that compact, quantized, domain-adapted LLMs can generate useful swarm BTs when embedded in a validated systems pipeline. They also show that parser acceptance and safety filtering remain necessary execution gates; generation quality alone is not sufficient for autonomous deployment.
\end{abstract}


\section{Introduction}
\label{sec:introduction}

Robotic swarms use the collective behavior of many relatively simple robots to perform tasks that are difficult, costly, or fragile for a single robot, including distributed exploration, monitoring, search, and cooperative transport \cite{brambilla2013swarm,navarro2013swarm}. Their strength is also a usability challenge: useful global behavior emerges from local sensing, coordination rules, communication assumptions, and execution constraints that are usually designed by specialists. Consequently, programming a swarm often requires explicit behavior design, simulator knowledge, and careful validation before deployment.

Large language models (LLMs) offer a promising interface between human intent and robot programs because they can interpret free-form commands and generate structured artifacts such as code, plans, and behavior trees \cite{naveed2024comprehensive,matarazzo2025survey}. Recent work has explored LLMs for multi-robot reasoning and swarm coordination \cite{strobel2024llm2swarm,li2025large}, and for behavior-tree (BT) generation in robotics \cite{lykov2023llmbrain,zhou2024llmbt,izzo2024btgenbot}. However, a practical swarm interface must satisfy requirements that are often studied separately: it should support non-expert language, generate executable multi-robot behaviors, run on realistic hardware budgets, and reject malformed, unsupported, or unsafe actions before execution.

This paper introduces \emph{CommandSwarm}, an end-to-end pipeline that converts multilingual speech or text commands into XML BTs for swarm control. CommandSwarm first translates or normalizes the user input into English, then checks the command with a safety classifier, constructs a constrained LLM prompt containing the permitted swarm primitives, generates an XML BT, and validates the output with a deterministic parser before execution in the Violet simulator. The design deliberately separates learned generation from execution authority: the LLM proposes a tree, while the parser and whitelist decide whether the tree is executable.

The study is organized around three research questions:
\begin{enumerate}[leftmargin=*,itemsep=2pt]
  \item \textbf{RQ1:} Which compact open LLMs are most reliable for XML BT generation under zero-shot, one-shot, and two-shot prompting?
  \item \textbf{RQ2:} Does parameter-efficient domain adaptation improve zero-shot BT generation beyond prompt engineering alone?
  \item \textbf{RQ3:} Which multilingual speech/text front-end provides the best quality-latency trade-off for natural-language swarm commands?
\end{enumerate}

The paper makes four contributions:
\begin{enumerate}[leftmargin=*,itemsep=2pt]
  \item \textbf{A safety-aware language-to-BT architecture for swarms.} We define a four-layer CommandSwarm pipeline covering translation, safety filtering, LLM-based BT synthesis, and middleware-level parsing/execution.
  \item \textbf{A systematic benchmark of compact open LLMs.} We evaluate eleven 6.7B--14B parameter models under zero-shot, one-shot, and two-shot prompting, with all models quantized to 4 bits for practical comparison.
  \item \textbf{A synthetic instruction--BT adaptation pipeline.} We construct a 2,063-example synthetic corpus and use LoRA to adapt Falcon3-Instruct-10B, substantially improving zero-shot generation quality and syntactic validity.
  \item \textbf{A multilingual front-end evaluation.} We compare Whisper and SeamlessM4T for speech translation, and EuroLLM variants for text translation, using BLEU, ROUGE-L, METEOR, and latency.
\end{enumerate}

The results show that prompt engineering alone can produce strong few-shot performance, but domain adaptation is decisive for robust zero-shot operation. More broadly, CommandSwarm demonstrates that reliable natural-language swarm control is a systems problem: translation, prompting, safety classification, parser validation, and action whitelisting must reinforce each other.

\section{Related Work}
\label{sec:related}

\subsection{Swarm Robotics and Human-Swarm Interaction}
Swarm robotics studies how groups of robots coordinate through local sensing, local communication, and decentralized control \cite{brambilla2013swarm,navarro2013swarm,shahzad2023swarm, majid2021lightweight}. Although swarm methods are attractive for robustness and scalability, they remain difficult to operate because high-level user intent must be mapped to coordinated low-level behaviors. Natural-language interaction has long been recognized as a way to reduce this barrier \cite{kemke2007saying}, but earlier systems lacked the language understanding and structured-generation capabilities needed for flexible behavior synthesis \cite{majid2023chirpy, majid2024challenging}.

LLMs have renewed interest in human-swarm interaction. LLM2Swarm explores both indirect and direct integration of LLMs with robot swarms, including controller synthesis and robot-level reasoning \cite{strobel2024llm2swarm}. Surveys of LLMs for multi-robot systems identify task allocation, planning, communication, and safety as key opportunities and open challenges \cite{li2025large, majid2024drl}. CommandSwarm is aligned with this direction but focuses on a narrower and more deployable objective: converting natural-language commands into parser-validated BTs for swarm scenarios using compact open models.

\subsection{Behavior Trees for Robot Task Representation}
Behavior trees provide a modular, reactive, and interpretable representation for robot behavior \cite{colledanchise2018bt,iovino2022survey}. A BT is composed of control-flow nodes, such as sequences and fallbacks, and execution nodes, such as actions and conditions. This structure is attractive for LLM-generated robot programs because the output can be inspected, parsed, and rejected if it violates the allowed syntax or action set.

Recent robotics work has used LLMs to generate BTs from text. LLM-BRAIn fine-tuned an Alpaca-style 7B model for BT generation \cite{lykov2023llmbrain}. LLM-BT uses LLM reasoning and BT updates for adaptive robotic tasks \cite{zhou2024llmbt}. BTGenBot demonstrated that lightweight 7B models can generate BTs for robotic tasks when fine-tuned and validated \cite{izzo2024btgenbot}; BTGenBot-2 later pushed this line toward smaller models, standardized benchmarking, and validator-based recovery \cite{izzo2026btgenbot2}. CommandSwarm differs in three ways: it targets swarm-level behaviors, evaluates a broader set of contemporary open LLMs under the same prompting protocol, and integrates multilingual input plus command-level safety filtering in the same pipeline.

\subsection{Multilingual Speech and Text Translation}
A natural interface for swarm control should not assume that operators type English commands. Whisper showed that large-scale weak supervision can support robust multilingual speech recognition and translation \cite{radford2022whisper}. SeamlessM4T provides a unified multilingual and multimodal translation model for speech and text tasks \cite{barrault2023seamless}. EuroLLM-9B was designed to improve open multilingual LLM support for European languages \cite{martins2025eurollm}. CommandSwarm evaluates these components as part of the end-to-end system rather than treating translation as an external preprocessing step.

\subsection{Safety Guardrails for LLM-Enabled Robots}
LLM-enabled robots require safeguards because natural-language commands can be ambiguous, unsupported, or unsafe. Llama Guard frames safety filtering as prompt/response classification using an explicit risk taxonomy \cite{inan2023llamaguard}. Robotics-specific work such as RoboGuard argues that safety rules should be grounded in robot context and checked before execution \cite{ravichandran2025roboguard}. CommandSwarm follows the same defense-in-depth principle at a lighter system level: the safety layer filters the command, while the parser rejects malformed XML and any node outside the approved action/condition whitelist. Unlike temporal-logic safety controllers, the present implementation does not yet prove runtime safety; it prevents unsupported BTs from reaching execution and leaves formal safety verification as future work.

\section{CommandSwarm Architecture and Methodology}
\label{sec:methodology}

\subsection{System Overview}
\begin{figure*}[!t]
  \centering
  \includegraphics[width=0.92\textwidth]{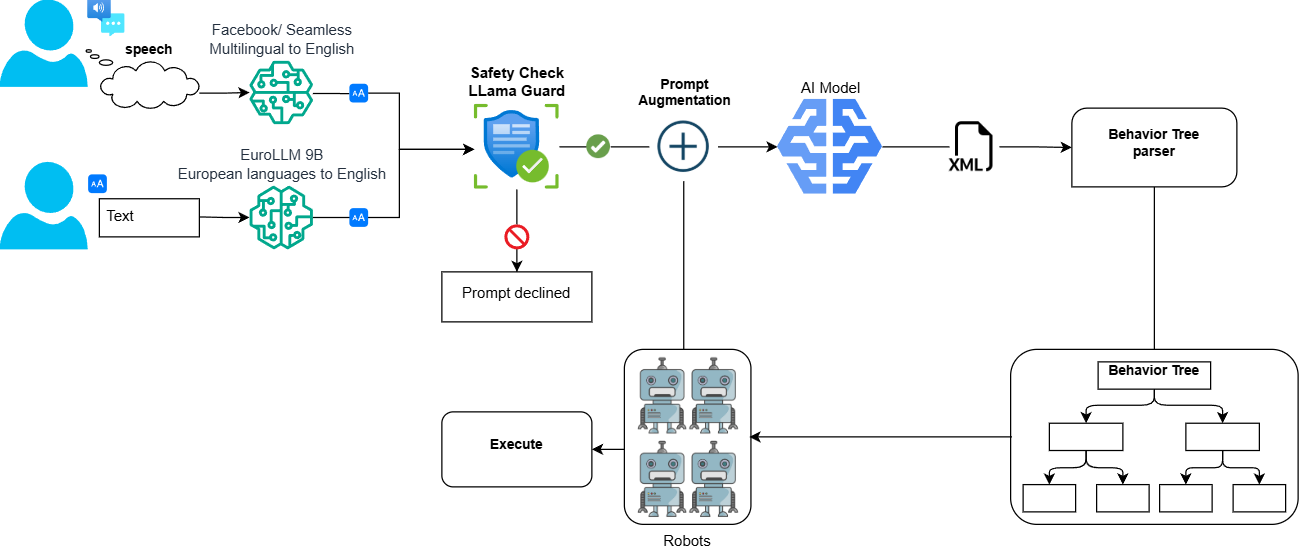}
  \caption{CommandSwarm system overview. User speech or text is translated into English, filtered for safety, converted by an LLM into an XML BT, and validated by the middleware parser before execution.}
  \label{fig:architecture}
\end{figure*}

Figure~\ref{fig:architecture} summarizes CommandSwarm. The pipeline receives a spoken or typed command, normalizes it into English, checks it for safety, generates an XML BT with an LLM, validates the tree against a closed set of swarm primitives, and passes the validated tree to the simulator or robot middleware.

\subsection{Translation Layer}
The translation layer supports two input modes. For speech, audio captured from a web microphone is resampled and processed by a speech translation model. We compare Whisper-medium and SeamlessM4T v2-large. For typed text, commands are translated or normalized with EuroLLM variants. The output of this layer is a single English command, which avoids requiring the downstream BT generator to handle all languages directly.

\subsection{Safety Layer}
Before generation, CommandSwarm passes the normalized command to a Llama-Guard-style safety classifier. Commands classified as unsafe or outside the intended operational domain are rejected before they reach the BT generator. This early filtering is complemented by parser-level validation after generation, creating two independent control points: one at the command level and one at the executable-plan level. In this paper, the safety layer is evaluated as part of the system design rather than as an independently benchmarked classifier; the experiments focus on translation quality, BT-generation quality, and parser acceptance.

\subsection{Behavior-Tree Generation Layer}
The LLM layer converts the safe English command into an XML BT. The prompt contains: (i) a system instruction defining the model role, (ii) the complete list of allowed action and condition nodes, (iii) the required XML skeleton, and (iv) the user command. The key design choice is to expose the action space explicitly. Rather than asking the model to invent robot behaviors, CommandSwarm asks it to arrange a fixed set of primitives into a valid BT.

\begin{lstlisting}[caption={Simplified CommandSwarm prompt structure for XML behavior-tree generation.},label={lst:prompt}]
SYSTEM: Generate only a valid XML behavior tree.
INSTRUCTIONS: Use only the listed actions and conditions.
REQUIRED FORMAT: <root> ... <BehaviorTree> ... </BehaviorTree>
                 <TreeNodesModel> ... </TreeNodesModel> </root>
USER COMMAND: <translated user command>
RESPONSE: XML only.
\end{lstlisting}

The generated tree must contain a single root, a BT element, and a \texttt{TreeNodesModel} declaration. Structural BT nodes such as \texttt{Sequence} and \texttt{Fallback} are allowed, while action and condition leaves must match the whitelist. Any explanatory text, malformed XML, missing root element, or unsupported node is treated as a failed generation.

\subsection{Middleware and Execution Layer}
The middleware parses the XML tree and maps its nodes to simulator or robot behaviors. The parser checks XML well-formedness, required attributes, root structure, and node legality. If any unsupported tag or malformed structure is found, execution is aborted and an error is returned. Valid trees are executed by the low-level control layer, where action nodes implement movement, signaling, obstacle avoidance, target search, or coordination behaviors in the Violet simulator.

\subsection{Synthetic Dataset and LoRA Adaptation}
\begin{figure}[!t]
  \centering
  \includegraphics[width=0.92\columnwidth]{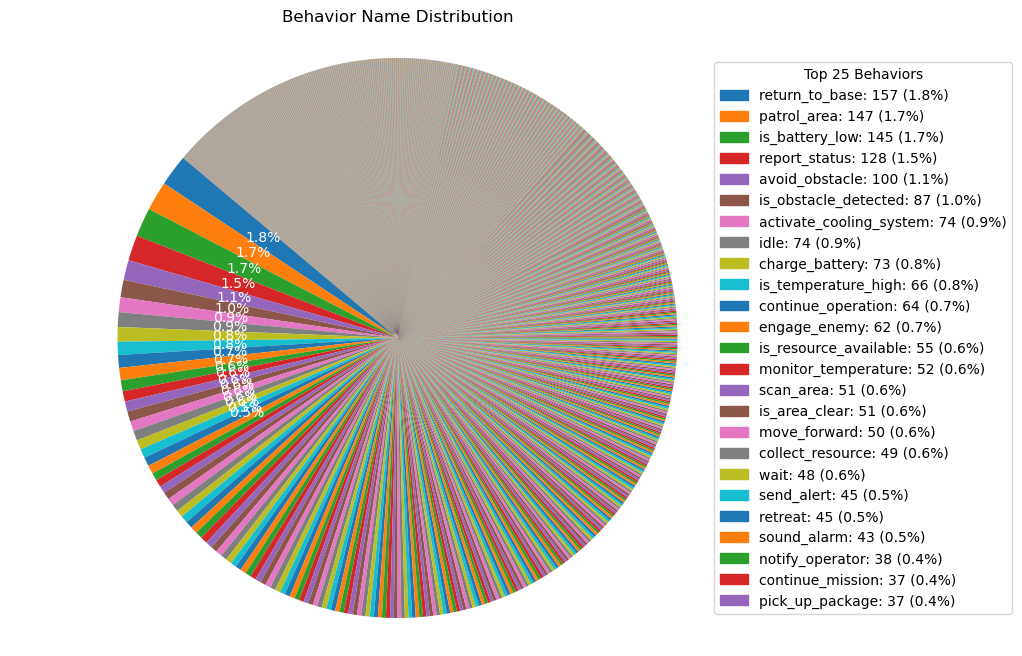}
  \caption{Distribution of behavior names in the synthetic instruction--BT dataset used for LoRA adaptation.}
  \label{fig:behavior_distribution}
\end{figure}

To improve domain reliability, we generated a synthetic instruction--BT dataset containing 2,063 pairs. The dataset covers diverse user commands and BT structures, including navigation, target search, obstacle handling, formation, signaling, and status reporting. Figure~\ref{fig:behavior_distribution} illustrates the distribution of behavior names in the synthetic corpus.

We fine-tuned the strongest selected model with Low-Rank Adaptation (LoRA), which freezes the base model and trains small low-rank adapter matrices \cite{hu2021lora}. In our implementation, LoRA adapters are applied to the attention projection modules. This parameter-efficient strategy is appropriate for CommandSwarm because the target task is narrow: the model does not need to learn general robotics from scratch, but to reliably express user intent using the permitted BT vocabulary and XML schema.

\subsection{Output Validation and Failure Categories}
Each generated output is passed through the same validation pipeline. We classify failures into four categories: non-XML responses, malformed XML, structurally incomplete BTs, and unsupported node names. Parser-accepted syntactic validity is therefore a stricter criterion than merely checking that the response contains XML-like text. This distinction is important because a BT can look plausible to a reader while still being non-executable by the middleware.

\section{Experimental Design}
\label{sec:experiments}

\subsection{Simulator and Task Scenarios}
Experiments were conducted with the Violet simulator, a lightweight PyGame-based swarm simulation environment \cite{violetdocs}. We use five canonical swarm scenarios that combine sensing, navigation, signaling, and coordination. Table~\ref{tab:scenarios} lists the task descriptions used for model screening.

\begin{table}[!t]
\centering
\caption{Swarm scenarios used for model screening.}
\label{tab:scenarios}
\footnotesize
\begin{tabularx}{\columnwidth}{cX}
\toprule
\textbf{ID} & \textbf{Scenario} \\
\midrule
1 & Detect an obstacle, avoid it, and change color to green. \\
2 & Wander until a target is detected, approach it, and signal achievement by changing color to red. \\
3 & Check whether the path is clear and form a line at the center. \\
4 & Find the goal, signal success by changing color to red, and align movement with other swarm agents. \\
5 & Detect the target and freeze movement after reaching it. \\
\bottomrule
\end{tabularx}
\end{table}

\subsection{Translation Evaluation}
We evaluate speech and text translation across nine European languages. For speech, we compare Whisper-medium and SeamlessM4T v2-large. For text, we compare EuroLLM-1.7B-Instruct and EuroLLM-9B-Instruct, both in 4-bit quantized form. Translation quality is measured with BLEU, ROUGE-L, and METEOR; latency is measured as average end-to-end inference time per command.

Whisper-medium reaches BLEU 0.50, ROUGE-L 0.71, and METEOR 0.75, with an average latency of approximately 5.2s per utterance. SeamlessM4T v2-large obtains BLEU 0.46, ROUGE-L 0.72, and METEOR 0.77, with a lower average latency of approximately 4.0s. For text translation, EuroLLM-1.7B performs poorly in this setup (BLEU 0.02, ROUGE-L 0.39, METEOR 0.43, and approximately 52s per sentence), while EuroLLM-9B achieves BLEU 0.62, ROUGE-L 0.81, METEOR 0.85, and approximately 5.5s per sentence.

\begin{figure*}[!t]
  \centering
  \begin{tabular}{cc}
    \includegraphics[width=0.46\textwidth]{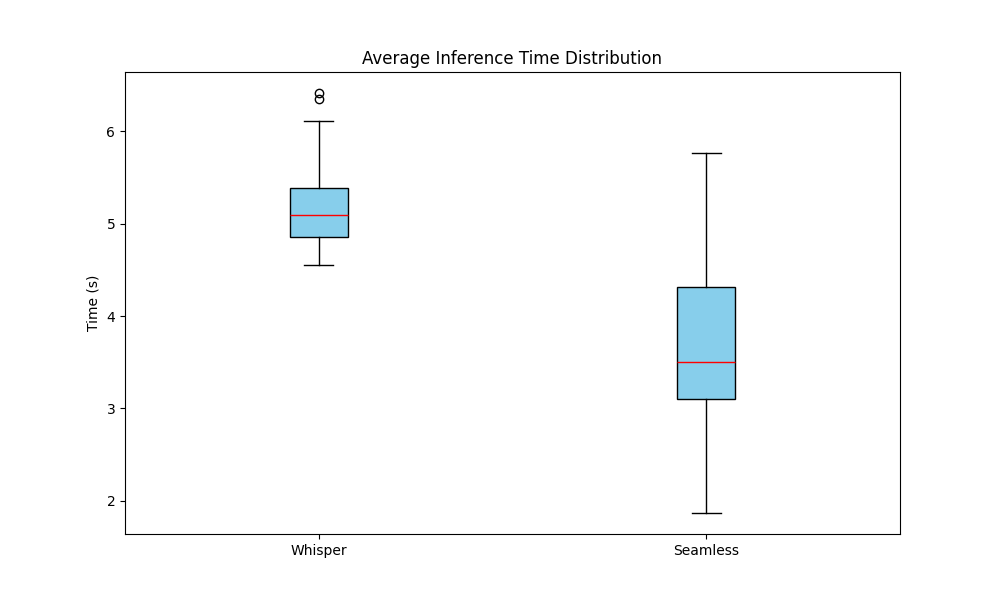} &
    \includegraphics[width=0.46\textwidth]{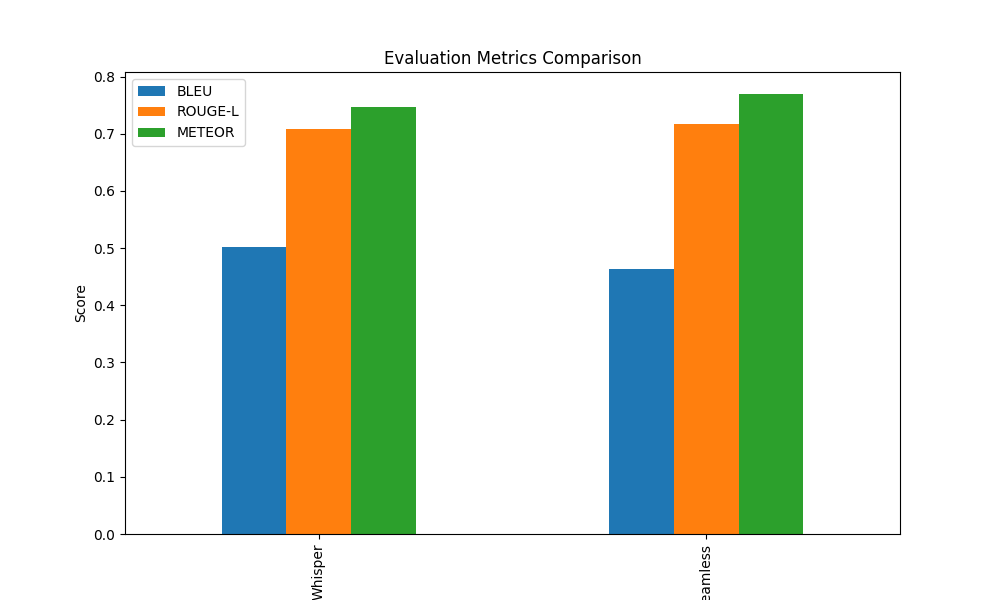} \\
    \includegraphics[width=0.46\textwidth]{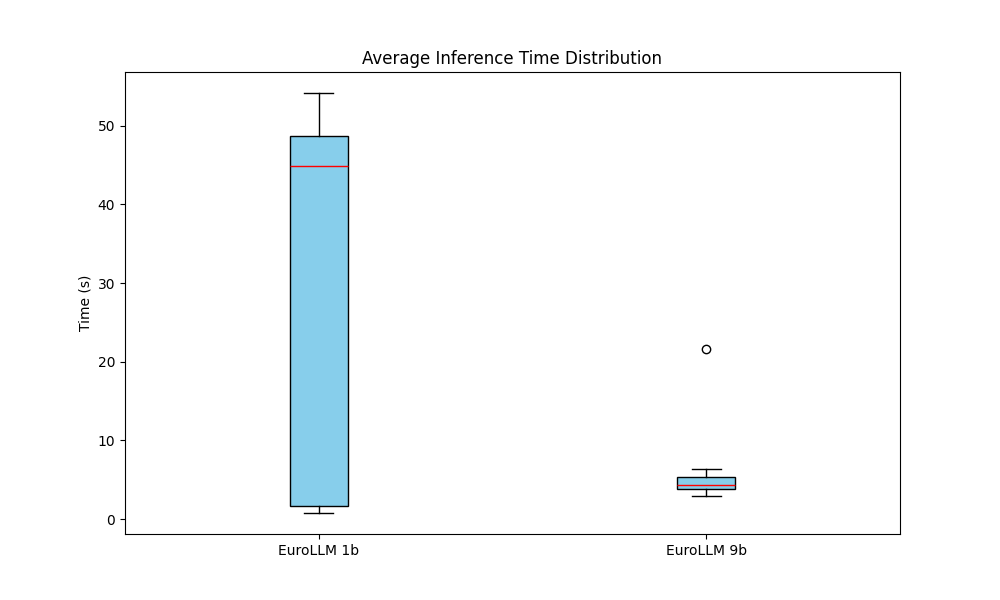} &
    \includegraphics[width=0.46\textwidth]{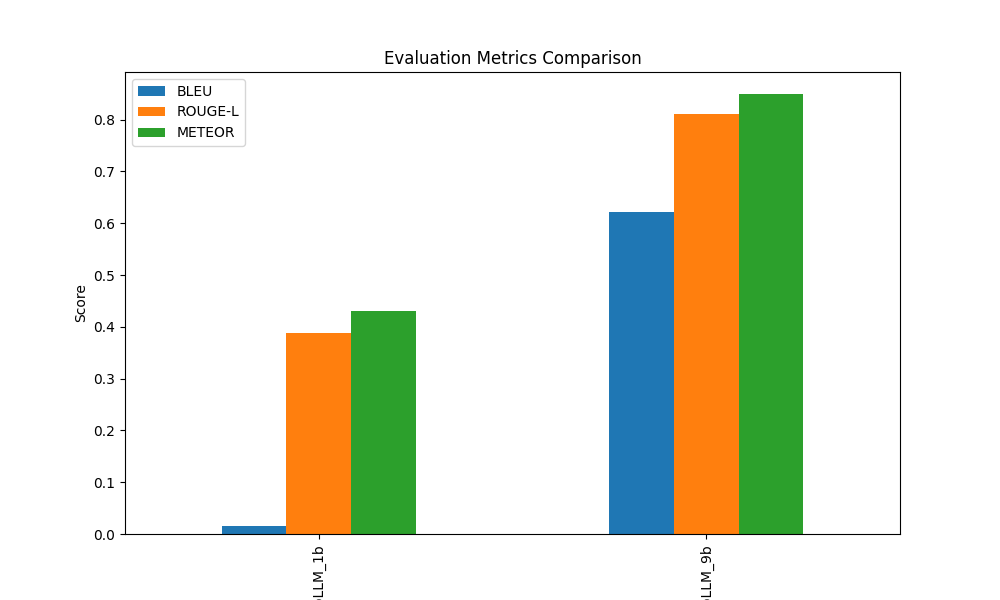}
  \end{tabular}
  \caption{Translation latency and quality. Top row: Whisper-medium versus SeamlessM4T v2-large for speech translation. Bottom row: EuroLLM-1.7B versus EuroLLM-9B for text translation.}
  \label{fig:translation_results}
\end{figure*}

\subsection{LLM Candidates for Behavior-Tree Generation}
We benchmark eleven open-source models, all evaluated with 4-bit quantization and the same prompting protocol. Table~\ref{tab:models} summarizes the candidate set.

\begin{table}[!t]
\centering
\caption{LLMs evaluated for behavior-tree generation.}
\label{tab:models}
\footnotesize
\begin{tabularx}{\columnwidth}{Xcc}
\toprule
\textbf{Model} & \textbf{Size} & \textbf{Precision} \\
\midrule
CodeLlama-7B \cite{roziere2024codellama} & 7B & 4-bit \\
DeepSeek-R1-Distill-Qwen-7B \cite{deepseekai2025deepseekr1} & 7B & 4-bit \\
DeepSeek-R1-Distill-Llama-8B \cite{deepseekai2025deepseekr1} & 8B & 4-bit \\
EuroLLM-Instruct-9B \cite{martins2025eurollm} & 9B & 4-bit \\
Falcon3-Instruct-10B \cite{tii2024falcon3} & 10B & 4-bit \\
Meta-Llama-3.1-Instruct-8B \cite{grattafiori2024llama3} & 8B & 4-bit \\
Mistral-7B-v3 \cite{jiang2023mistral} & 7B & 4-bit \\
Qwen2.5-Coder-Instruct-7B \cite{hui2024qwen25coder} & 7B & 4-bit \\
DeepSeek-Coder-Base-7B \cite{guo2024deepseekcoder} & 7B & 4-bit \\
DeepSeek-Coder-Instruct-6.7B \cite{guo2024deepseekcoder} & 6.7B & 4-bit \\
Phi-4-14B \cite{abdin2024phi4} & 14B & 4-bit \\
\bottomrule
\end{tabularx}
\end{table}

\subsection{Prompting Conditions and Metrics}
Each model is evaluated under three prompting conditions: zero-shot, one-shot, and two-shot. In zero-shot prompting, the model receives only the system instructions, allowed behavior list, XML format constraints, and target command. In one-shot and two-shot prompting, one or two complete instruction--BT examples are prepended before the target command.

We report three metrics. \emph{BLEU} measures n-gram overlap between the generated BT and a reference tree. \emph{ROUGE-L} measures longest-common-subsequence similarity. \emph{Parser-accepted syntactic correctness} is the percentage of outputs that are well-formed XML, have the required BT structure, and use only whitelisted nodes. BLEU and ROUGE-L are useful for measuring structural similarity to the reference, but they are not sufficient indicators of semantic correctness; therefore, parser acceptance is treated as a necessary execution gate.

\subsection{Two-Stage Model Selection}
The first stage evaluates all eleven models on the five scenario descriptions in Table~\ref{tab:scenarios}. This screening stage is intentionally small and is used to identify promising models, not to make final performance claims. The second stage focuses on the three strongest models from stage one: Falcon3-Instruct-10B, Mistral-7B-v3, and Qwen2.5-Coder-Instruct-7B. These models are evaluated on a larger held-out set of 50 examples before selecting Falcon3-Instruct-10B for LoRA adaptation.

\section{Results}
\label{sec:results}

\subsection{Stage-One Model Screening}
\begin{figure*}[!t]
  \centering
  \begin{tabular}{@{}ccc@{}}
    \includegraphics[width=0.305\textwidth]{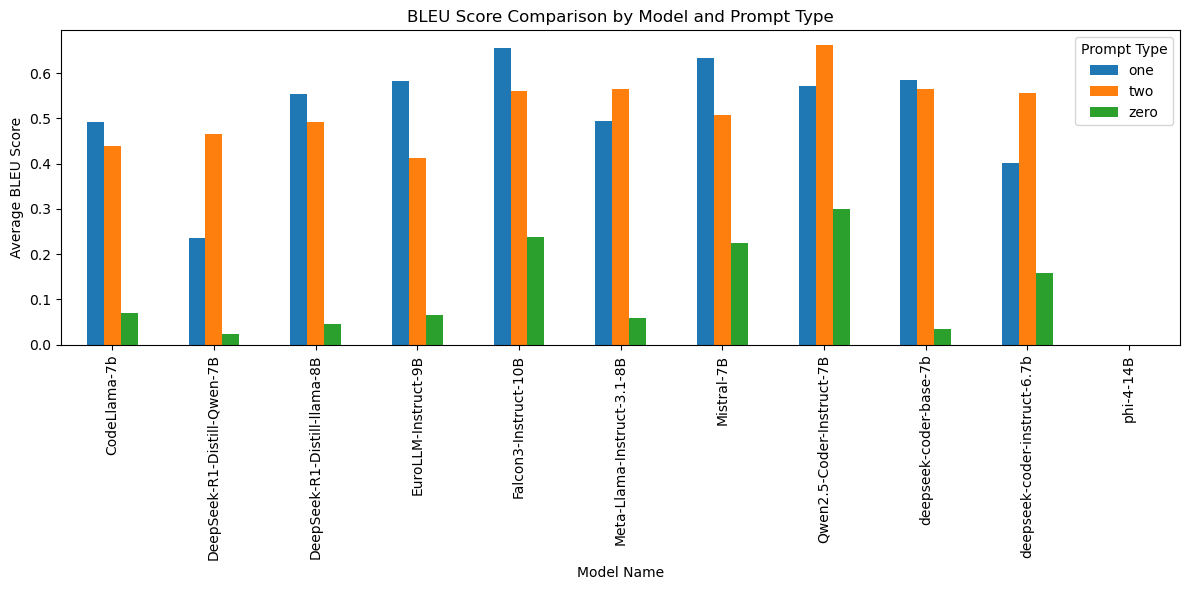} &
    \includegraphics[width=0.305\textwidth]{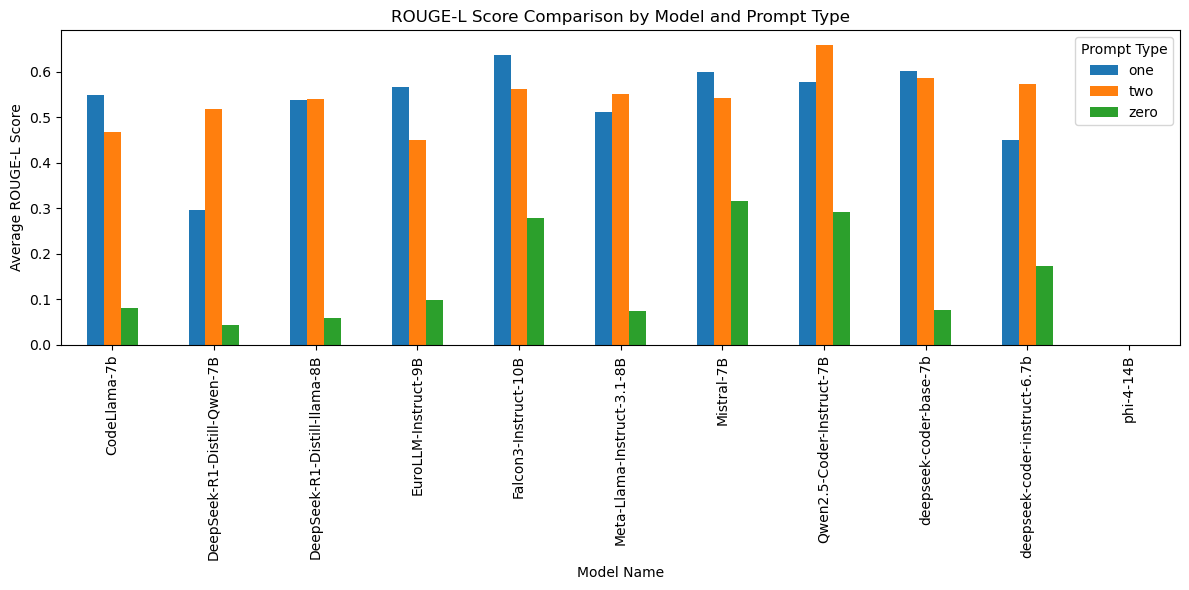} &
    \includegraphics[width=0.305\textwidth]{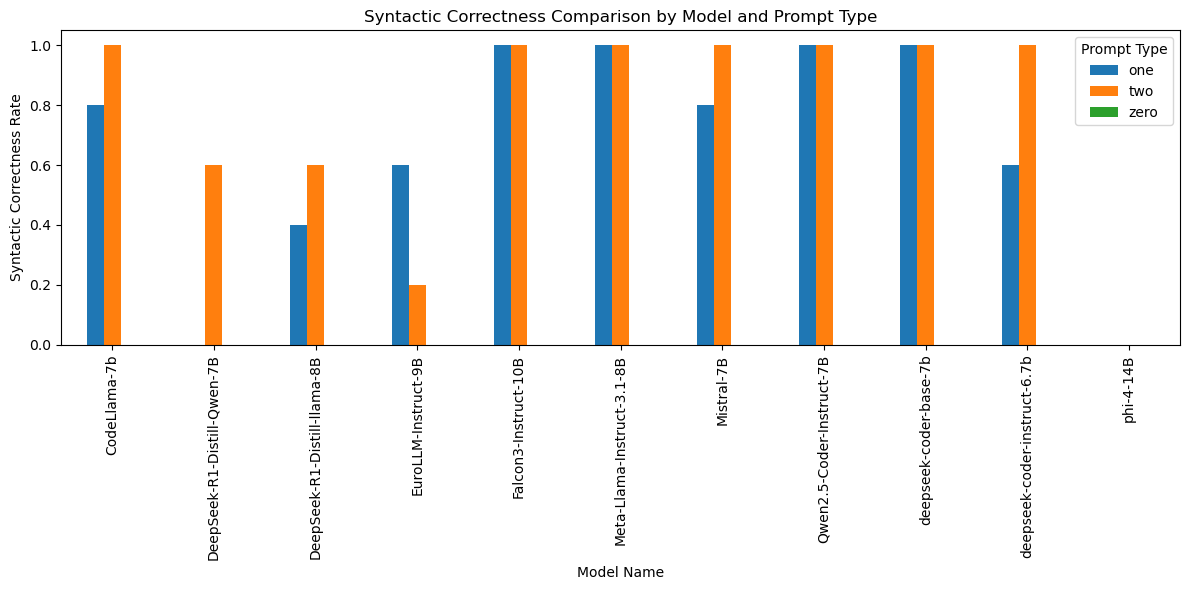}
  \end{tabular}
  \caption{Stage-one comparison of eleven 4-bit quantized LLMs under zero-shot, one-shot, and two-shot prompting. Left: BLEU. Middle: ROUGE-L. Right: syntactic correctness.}
  \label{fig:model_selection}
\end{figure*}

The first screening stage shows that model choice and prompt design have a major effect on BT generation quality. Figure~\ref{fig:model_selection} compares BLEU, ROUGE-L, and syntactic correctness for all eleven models across zero-shot, one-shot, and two-shot prompting. Qwen2.5-Coder-Instruct-7B, Falcon3-Instruct-10B, and Mistral-7B-v3 consistently rank among the strongest models. Few-shot examples are especially important for syntax: several models approach or reach full syntactic validity in one-shot or two-shot settings, whereas zero-shot prompting remains unreliable for many candidates.

\subsection{Stage-Two Evaluation of the Top Models}
\begin{figure*}[!t]
  \centering
  \begin{tabular}{@{}ccc@{}}
    \includegraphics[width=0.305\textwidth]{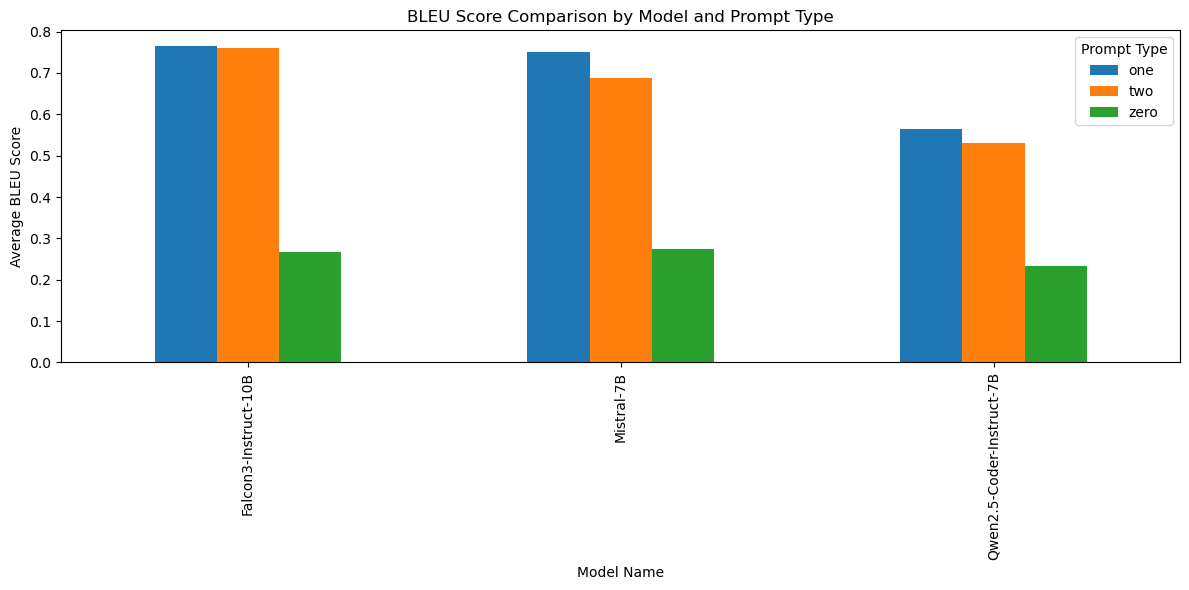} &
    \includegraphics[width=0.305\textwidth]{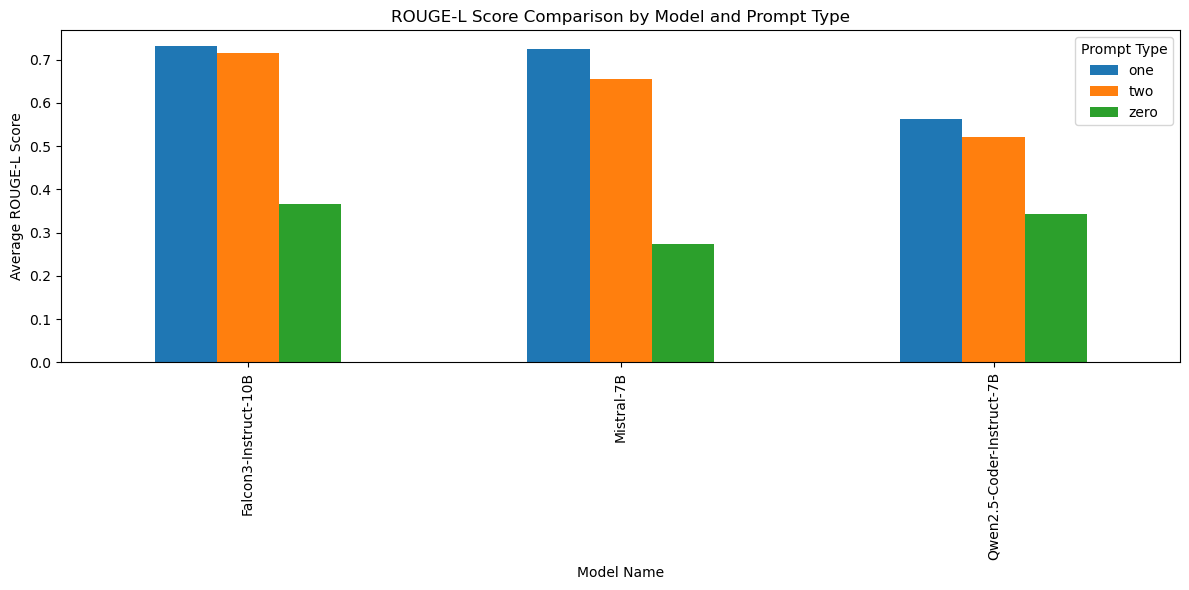} &
    \includegraphics[width=0.305\textwidth]{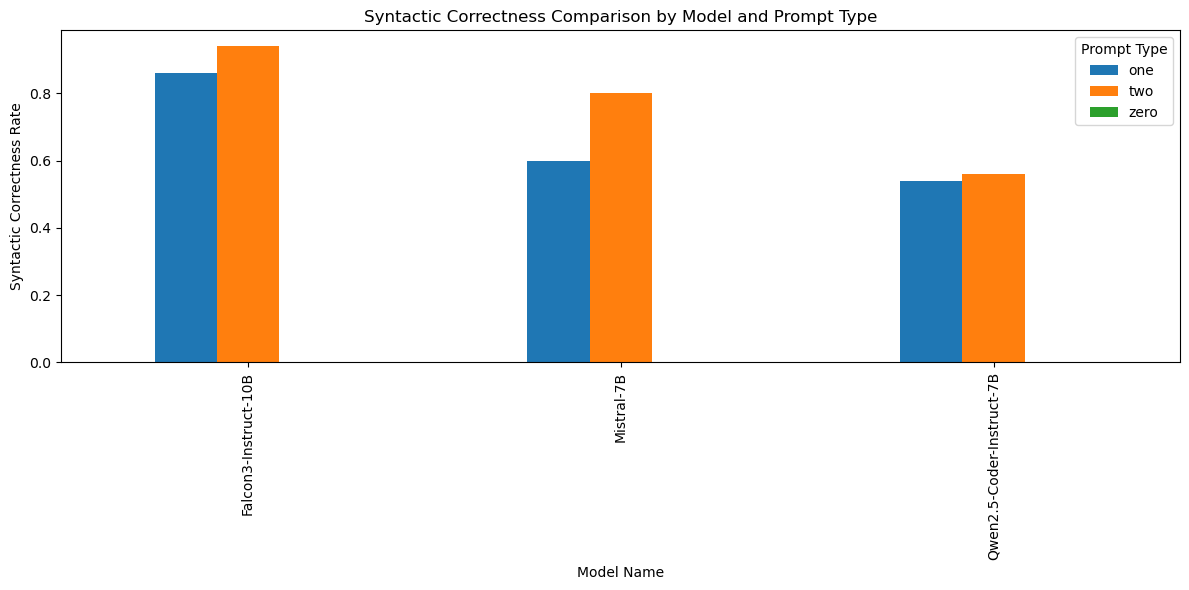}
  \end{tabular}
  \caption{Stage-two evaluation of the strongest three LLMs on 50 held-out behavior-tree examples. Left: BLEU. Middle: ROUGE-L. Right: syntactic correctness.}
  \label{fig:top3}
\end{figure*}

The second stage evaluates the top three candidates on 50 held-out examples. Figure~\ref{fig:top3} shows that Falcon3-Instruct-10B and Mistral-7B-v3 retain the strongest overall performance, while Qwen2.5-Coder-Instruct-7B remains competitive but less consistent. Two-shot prompting provides the highest syntactic reliability, but zero-shot performance still leaves substantial room for improvement. This motivates parameter-efficient fine-tuning rather than relying only on increasingly long prompts.

\subsection{Effect of LoRA Fine-Tuning}
\begin{figure*}[!t]
  \centering
  \begin{tabular}{@{}ccc@{}}
    \includegraphics[width=0.305\textwidth]{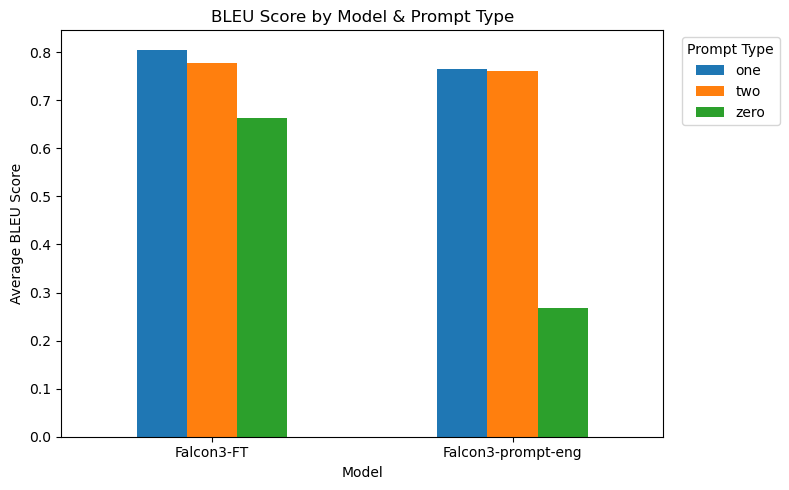} &
    \includegraphics[width=0.305\textwidth]{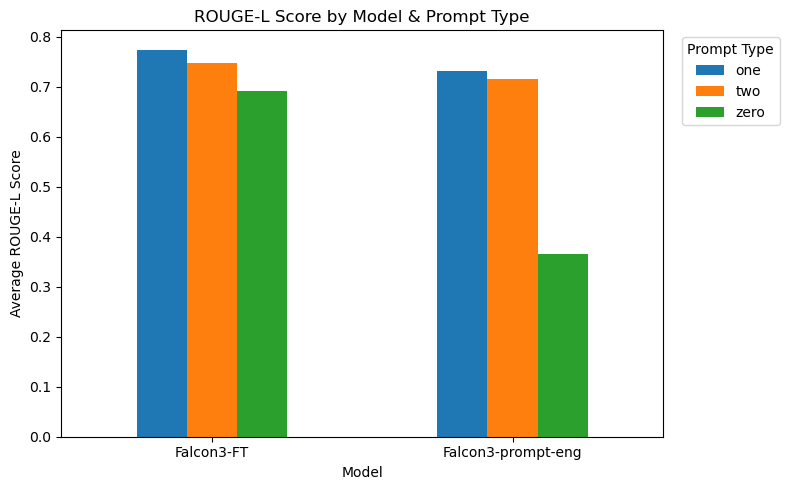} &
    \includegraphics[width=0.305\textwidth]{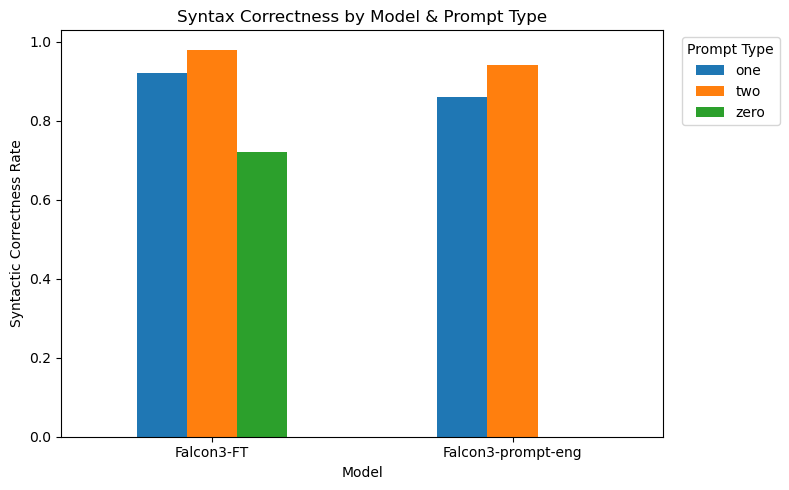}
  \end{tabular}
  \caption{Prompt-engineered Falcon3 versus LoRA-adapted Falcon3-FT on 50 held-out examples. Left: BLEU. Middle: ROUGE-L. Right: syntactic correctness.}
  \label{fig:finetuning}
\end{figure*}

We compare the fine-tuned Falcon3 model (Falcon3-FT) with the best prompt-engineered Falcon3 baseline on the same 50-example held-out set. Figure~\ref{fig:finetuning} and Table~\ref{tab:finetuning_summary} summarize the results. Falcon3-FT improves BLEU in all prompting settings: one-shot BLEU increases from 0.765 to 0.805, two-shot BLEU from 0.761 to 0.777, and zero-shot BLEU from 0.267 to 0.663. The zero-shot gain of 0.396 is the largest improvement.

ROUGE-L follows the same trend. One-shot ROUGE-L increases from 0.731 to 0.777, two-shot from 0.715 to 0.748, and zero-shot from 0.366 to 0.692. Parser-accepted syntactic correctness also improves: one-shot validity rises from 86\% to 92\%, two-shot from 94\% to 98\%, and zero-shot from 0\% to 72\%. These results indicate that the LoRA-adapted model internalizes much of the BT schema and action vocabulary that otherwise must be provided through examples.

\begin{table}[!t]
\centering
\caption{Falcon3 baseline versus LoRA-adapted Falcon3-FT on the held-out 50-example set.}
\label{tab:finetuning_summary}
\footnotesize
\begin{tabular}{lccc}
\toprule
\textbf{Setting} & \textbf{Metric} & \textbf{Baseline} & \textbf{Falcon3-FT} \\
\midrule
Zero-shot & BLEU & 0.267 & 0.663 \\
Zero-shot & ROUGE-L & 0.366 & 0.692 \\
Zero-shot & Syntax & 0\% & 72\% \\
\midrule
One-shot & BLEU & 0.765 & 0.805 \\
One-shot & ROUGE-L & 0.731 & 0.777 \\
One-shot & Syntax & 86\% & 92\% \\
\midrule
Two-shot & BLEU & 0.761 & 0.777 \\
Two-shot & ROUGE-L & 0.715 & 0.748 \\
Two-shot & Syntax & 94\% & 98\% \\
\bottomrule
\end{tabular}
\end{table}

\subsection{Translation Results}
The translation experiments show a clear separation between model variants. SeamlessM4T v2-large provides the strongest speech-translation trade-off in this setup: it is faster than Whisper-medium and achieves slightly higher ROUGE-L and METEOR, despite lower BLEU. For text translation, EuroLLM-9B substantially outperforms EuroLLM-1.7B in both quality and latency. These results justify the final CommandSwarm configuration: SeamlessM4T for spoken commands and EuroLLM-9B for typed multilingual input.

\subsection{Overall System Interpretation}
The combined results support three conclusions. First, compact open models can generate useful BTs, but the best model must be selected empirically. Second, few-shot prompting is effective for enforcing XML form, yet it does not fully solve zero-shot brittleness. Third, lightweight fine-tuning is highly effective when the target output language is narrow and structured. The best CommandSwarm configuration, therefore, combines a compact quantized LLM, constrained prompting, LoRA adaptation, and deterministic parser validation.

\section{Discussion}
\label{sec:discussion}

\subsection{Answering the Research Questions}
For RQ1, the results show that Falcon3-Instruct-10B and Mistral-7B-v3 are the strongest prompt-engineered candidates among the evaluated compact open models. Qwen2.5-Coder-Instruct-7B is competitive, but less consistent across prompting settings. For RQ2, LoRA adaptation clearly improves zero-shot behavior: the largest gains occur when no examples are provided at inference time. For RQ3, SeamlessM4T v2-large provides the strongest speech front-end trade-off in this setup, while EuroLLM-9B is the best text-translation option among the evaluated variants.

\subsection{Why Fine-Tuning Helps}
The largest improvements appear in zero-shot generation. This is important for real use because users should not need to provide examples when commanding a swarm. Prompt engineering can describe the XML schema and action vocabulary, but the model may still omit required tags, invent unsupported actions, or produce explanatory text. LoRA adaptation exposes the model repeatedly to the target representation, making the XML structure and BT vocabulary part of the model's task behavior rather than only part of the prompt context.

\subsection{Prompting and Validation Are Complementary}
The results should not be interpreted as replacing validation with fine-tuning. Even after adaptation, 28\% of zero-shot outputs fail parser validation. This confirms that LLM outputs must remain proposals, not executable authority. CommandSwarm therefore uses a defense-in-depth design: prompting reduces errors, LoRA adaptation improves the distribution of generated trees, the safety classifier filters commands, and the parser enforces the final execution boundary.

\subsection{Threats to Validity}
This study has four main threats to validity. First, BLEU and ROUGE-L measure similarity to reference XML but do not fully capture whether an alternative BT would achieve the same behavior. Second, parser acceptance is necessary for execution but not sufficient for semantic correctness in all environments. Third, the experiments are conducted in simulation; physical swarms introduce sensing noise, communication loss, battery constraints, and timing issues. Fourth, the safety classifier is integrated into the architecture but not independently evaluated against adversarial or domain-specific unsafe commands in this paper. These limitations mean that the results support CommandSwarm as a validated generation pipeline for controlled swarm scenarios, not as a complete safety assurance case for real-world autonomy.

\subsection{Future Work}
Future work will focus on three directions. The first is semantic validation, including learned or symbolic validators that check whether a generated BT satisfies the user intent beyond XML syntax. The second is simulator-in-the-loop evaluation, where generated BTs are executed and scored according to task success rather than only text similarity and parser validity. The third is real-robot transfer, starting with small Crazyflie or ePuck swarms to measure latency, communication overhead, and robustness under physical constraints. A fourth direction is hierarchical BT generation, where complex commands are decomposed into validated subtrees to overcome context-window limits and improve reuse.

\section{Safety and Ethical Considerations}
\label{sec:ethics}

Natural-language swarm control creates a direct path from user intent to coordinated robot behavior. This raises safety, accountability, privacy, and dual-use concerns. CommandSwarm addresses these concerns through layered restrictions rather than relying on the LLM alone. The safety classifier rejects unsafe or out-of-domain commands before generation. The prompt restricts the output to approved behavior primitives. The parser then rejects malformed XML or unlisted nodes before execution.

This design improves accountability because every accepted command can be logged together with the translated text, generated BT, parser result, and execution status. Such logs support debugging, auditability, and post-incident analysis. For privacy, audio and text commands should be processed locally when possible, minimized in logs, and stored only under explicit retention policies. For accessibility, multilingual support should be evaluated across accents, languages, and noisy environments rather than only on clean benchmark examples.

The current implementation is intended for research, education, and controlled simulation or laboratory environments. Real-world deployment should require additional safeguards, including operator approval for sensitive tasks, geofencing, emergency stop mechanisms, runtime monitors, and domain-specific risk assessment. The main ethical lesson is that LLM-enabled autonomy should be designed as a constrained toolchain: the language model can help translate intent, but deterministic safety gates must decide what is executable.

\section{Conclusion}
\label{sec:conclusion}

This paper presented CommandSwarm, a safety-aware pipeline for translating multilingual speech or text commands into executable XML BTs for robotic swarms. The system combines translation, command-level safety filtering, constrained LLM prompting, LoRA adaptation, and parser-level validation. Across eleven compact open LLMs, Falcon3-Instruct-10B and Mistral-7B-v3 achieved the strongest few-shot BT generation performance. Fine-tuning Falcon3-Instruct-10B with LoRA on 2,063 synthetic instruction--BT pairs produced the largest benefit in the zero-shot setting, raising BLEU from 0.267 to 0.663 and syntactic validity from 0\% to 72\%.

The central finding is that natural-language swarm control becomes substantially more reliable when generation is treated as one component in a validated systems pipeline. Model choice, prompting, domain adaptation, translation, safety filtering, and parser validation all contribute to the final behavior. CommandSwarm, therefore, offers a practical step toward accessible human-swarm interaction while preserving a clear boundary between language-model generation and executable robot behavior.

\bibliographystyle{IEEEtran}
\bibliography{mybibliography}

\end{document}